# Learning Representations from Persian Handwriting for Offline Signature Verification, a Deep Transfer Learning Approach

Omid Mersa
Department of Electrical and
Computer Engineering
University of Tehran
Tehran, Iran
omidme@ymail.com

Farhood Etaati
Department of Electrical and
Computer Engineering
University of Tehran
Tehran, Iran
farhoodetaati@gmail.com

Saeed Masoudnia
Department of Electrical and
Computer Engineering
University of Tehran
Tehran, Iran
masoudnia@ut.ac.ir

Babak Nadjar Araabi
Department of Electrical and
Computer Engineering
University of Tehran
Tehran, Iran
araabi@ut.ac.ir

*Abstract*— Offline Signature Verification (OSV) is a challenging pattern recognition task, especially when it is expected to generalize well on the skilled forgeries that are not available during the training. Its challenges also include small training sample and large intra-class variations. Considering the limitations, we suggest a novel transfer learning approach from Persian handwriting domain to multi-language OSV domain. We train two Residual CNNs on the source domain separately based on two different tasks of word classification and writer identification. Since identifying a person's signature resembles identifying one's handwriting, it seems perfectly convenient to use handwriting for the feature learning phase. The learned representation on the more varied and plentiful handwriting dataset can compensate for the lack of training data in the original task, i.e. OSV, without sacrificing the generalizability. Our proposed OSV system includes two steps: learning representation and verification of the input signature. For the first step, the signature images are fed into the trained Residual CNNs. The output representations are then used to train SVMs for the verification. We test our OSV system on three different signature datasets, including MCYT (a Spanish signature dataset), UTSig (a Persian one) and GPDS-Synthetic (an artificial dataset). On UT-SIG, we achieved 9.80% Equal Error Rate (EER) which showed substantial improvement over the best EER in the literature, 17.45%. Our proposed method surpassed state-of-the-arts by 6% on GPDS-Synthetic, achieving 6.81%. On MCYT, EER of 3.98% was obtained which is comparable to the best previously reported results.

*Keywords: Transfer learning, Persian Handwriting, Offline Signature Verification, Convolutional Network.*

## I. INTRODUCTION

Given handwritten signatures significance as one of the main methods of document authentication, the task of verifying signatures validity becomes of great importance. Offline Signature Verification (OSV) systems aim to distinguish whether a given signature image is produced by the claimed author (genuine) or by an impostor (forgery). The basic challenge in OSV compared to the other physical biometrics such as fingerprint or iris is having a high intra-class variation. The OSV problem becomes more difficult in the presence of skilled forgeries, where an impostor carefully attempts to forge a signature. The challenge is further aggravated in the real scenario of OSV when a few genuine signatures and no skilled forgeries are available for training.

The most previous researches on the OSV has been dedicated to designing hand-crafted feature extraction. However, In recent years, automatic feature learning by CNNs has significantly improved the performance for OSV [1, 2]. The current trend of CNNs literature suggests using deeper models for feature learning which could be beneficial to both performance of the system, and generalizability of the learned representations of training data. However, deeper networks need rich and plentiful training data, which is rare in signature datasets. In fact, collecting signature data is hard, expensive, and is riddled with security concerns.

Considering the limitations, utilizing a different training dataset for the feature learning, a set similar to signatures, but is greater in size and richness, might seem reasonable. Using a more varied, and plentiful dataset, trains the deep networks to learn better representations of images alike to signatures, which can compensate for the lack of training data in the original task, without sacrificing the generalizability and depth of the networks. The goal of transfer learning is to leverage knowledge from a source task to improve learning in a target task, mostly in cases when the two source and target tasks are similar. Since identifying a person's signature closely resembles identifying one's handwriting, it seems perfectly convenient to use handwriting data as the training input for the feature learning phase. Furthermore, gathering handwriting data is more economical, since it is more available, and not as personally critical as signature data. Therefore, collecting handwriting data is straightforward in both the number of samples and classes, which consequently gives us the chance to use deeper networks in order to extract more meaningful features of the learned data.

While transfer learning has been used in similar problems with great success, its utilization in the OSV domain is limited. In this paper, we investigate if using a deep model pretrained on a Persian handwriting dataset and transferring it to OSV result in a performance boost of OSV system.

In the following, we first review the related works. Our proposed method is then presented in section III. We finally test the proposed method on three OSV datasets and discuss the achieved results.

## II. RELATED WORKS

### A. Offline Signature Verification

Here we briefly review the main concepts of OSV and the recently published works. For a more detailed review of OSV see the recent review paper[1]. The most previous researches on the OSV has been dedicated to the design of hand-crafted feature extraction [1, 3-5]. However, In the last five years, automatic feature learning by CNNs has significantly improved the performance of OSV systems[1, 2]. Recently published researches [6, 7] have addressed the practical goal of OSV literature, i.e., verification of genuine signatures versus skilled forgeries. Most studies did not use skilled forgeries for training [7]. This scenario is the most general case in practical applications. However, Hafemann et al. [6] suggested using skilled forgeries of several users for training, but testing in a separate set of users. They proposed different formulations for OSV based on using both genuine signatures and skilled forgeries, via using a transfer learning model, where a source signature dataset, in this case, the GPDS960[8] dataset, was utilized to train a model, which would be further used as a feature extraction system aimed for a target signature data. However, collecting rich signature datasets is an arduous task which leads to most signature datasets being small-sized, besides the fact that in real-world situations there are strict limitations on the number of training samples, training OSV models solely on signature data seems like an impractical choice.

### B. Transfer Learning

Transfer learning is one of the most prominent approaches in the current deep learning literature, since training models from scratch is not usually a viable option and re-using pretrained models might also result in performance boosts. Some studies [9] [10] [11] developed different cross-domain transfer learning approaches by employing rich labels from the text domains in order to mitigate the problem of insufficient image training data. Other works [12] [13] [14] employed transfer learning for handwriting problems.

Transfer learning has been used in OSV problems, although limitedly. Alvarez et al.[15] employed a transfer learning approach by using a pretrained network and further fine-tuning its latter layers in an OSV problem. This network which was previously trained on the task of classification of ImageNet dataset, did remarkably well. This indicates the problem-solving potential of this approach, since the source domain (image classification) is quite dissimilar to the target domain (OSV). Furthermore, Hafemann et al.[6] proposed a two-phase pipeline which utilized transfer learning in order to verify signatures credibility. The suggested model employed classifiers each trained for a specific signer based on the features extracted by the convolutional network. In order to learn the most generalizable features possible, the CNN was trained on the identification task of the GPDS960 dataset, which is the biggest English signature dataset that is no longer available due to the recent EU's data protection regulations.

Table I  A Sample pool from the Persian handwriting dataset that we have used in this research.

| Words | Writer #1 | Writer #2 | Writer #1 |
|---|---|---|---|
| **Rayaneh (Computer)** | 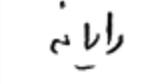 | 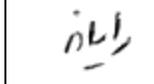 | 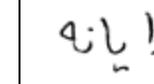 |
| **Farvardin (a Persian Month)** | 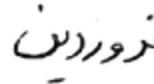 | 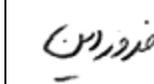 | 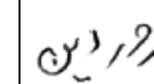 |

## III. PROPOSED METHOD

Our proposed OSV method consists of two phases: feature learning, and verification. In the first step of the feature learning phase, a CNN is trained on a source task in the handwriting domain. This task could be either of word recognition or writer identification, both on the source handwriting dataset. Next, by utilizing a transfer learning approach, the model will be transferred into the signature domain. Finally, for the verification phase, SVM classifiers will be trained for the task of user-dependent verification of signature images.

### A. Feature Generation using Transfer Learning

Training deep networks requires significant amounts of data, and if this constraint is not met, networks suffer from overfitting. Since handwritten signature datasets usually lack in quantity, the task of training a deep network based on this kind of data becomes even more challenging. To overcome this limitation, our proposed system uses a transfer learning paradigm and is built on ResNet CNNs which acts as the feature extractors as presented in [16]. Deep residual networks were first introduced by He et al.[16] as a mean to solve the issue of vanishing gradients in deep networks. This was achieved by adding the input of ResNet blocks to the outputs of such blocks through identity connections. Since the approach is based on transfer learning, we will illustrate the source and target domains, and our transfer learning strategies in the following lines.

▪ *Source Domain: Persian Handwriting Dataset*: We used a Persian handwriting dataset [17] as a source domain. This dataset consists of 115 words, where each was written and labeled differently for more than 500 different writers. Table 1 shows a sample pool of this dataset. By utilizing this type of data as a source domain which closely resembles the target domain, we have to define domain-specific tasks which learn translate-able features for the signature domain. We use this by selecting two different tasks on the Persian handwriting dataset, Word Recognition, and Writer Identification.

▪ *Target Domain, Handwritten Signatures:* The goal of this project is to solve the writer-dependent OSV. Since signatures vary heavily between cultures, the task of creating a general signature verifier which is not biased towards any kind of signature data is more challenging than type-specific signature verifiers. Since Persian signatures and Latin signatures offer a good contrast, creating an OSV model which handles both types of data well, is of great importance in this research.

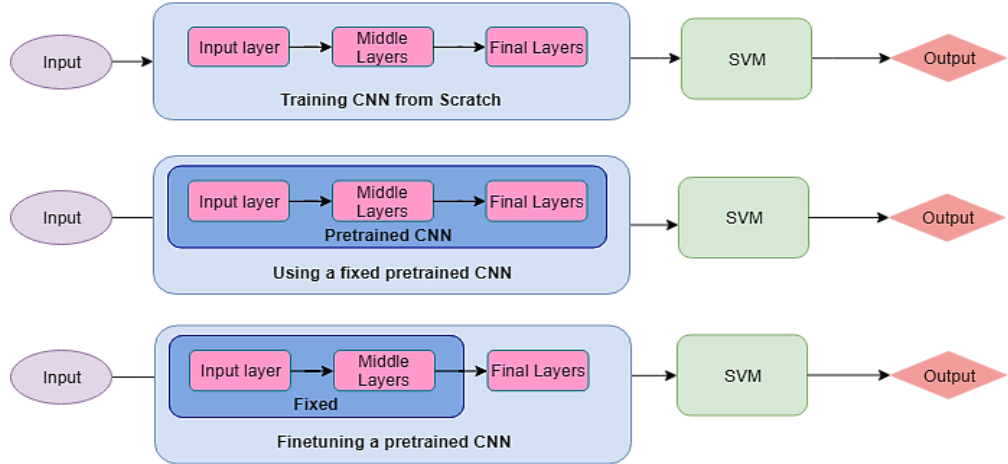

Fig. 1  Three learning strategies for the feature extraction network: *Training from scratch, transferring a fixed pretrained model, and finetuning a pretrained model on the destination task*

▪ ***Training process of the feature generator:*** Before further detailing the process, we first denote the learning algorithm. We can train the network using three approaches. Learning the CNN from scratch, utilizing a CNN trained on a source task with fixed parameters, fine-tuning a CNN trained on a source task by training the network on the final task. An overview of the mentioned learning methods is presented in *Fig. 1*.

## B. Signature Verification

After learning good representations of the training data, a classifier will be trained for each signatory. These classifiers will be trained on the features extracted by the deep network from the input data. In the training process, we only use random forgeries which will be generated from randomly picking genuine signatures of other signatories, as the negative samples, since acquiring skilled forgery samples is not plausible in the real situations. Since SVMs[18] usually fare better in generalization and do not require much training data, SVMs were chosen as classifiers for this task. As it is thoroughly explained in [18], support vector machines aim to find hyperplanes that have the largest distances from any training data, from any class. This feat is done by maximizing Equation 1.

$$\max L_d = \sum a_i - \frac{1}{2}\sum a_i a_j y_i y_j k(x_i . x_j), \qquad s.t \sum_{i=1}^{n} a_i y_i = 0 \quad (1)$$

Where $k(x, y)$ is the kernel function. The kernel functions used in this study are described in TABLE II.

## C. Summary of Our Method

Our proposed OSV includes two steps: feature generation, and the final verification. In the first step, we use a CNN (ResNet-8) which was previously trained on tasks of writer identification and word recognition on a Persian handwriting dataset. For the verification step, we use an SVM classifier for each user to separate the genuine samples of the user from other users' genuine samples (known as random forgeries) in the training phase. In short, the proposed method can be summarized in the following pseudo-code.

Pseudo-code for training our proposed method
1. Preprocessing of the dataset.
2. Dividing the datasets into train and test data subsets.
3. Training of the feature generator network
   a. Training the feature generator network on a source task, e.g.
   b. Testing the trained network in order to analyze its performance.
   c. Transferring the networks weights to a target task, by either fine-tuning or not.
4. Training of the Classifiers:
   a. Generate an SVM for each user class.
   b. Generate a set of random forgeries for each signature class.
   c. Train the SVM on the given genuine signatures of the a particular user and random forgery signatures of others.

## IV.   EXPERIMENTAL RESULTS

Several experiments were conducted to thoroughly analyze the performance of our transfer learning approach for the problem of OSV. A Persian handwriting dataset was used as the source domain. The signature datasets as the target domain in this study are MCYT-75 [19], UTSig [20] and GPDS-synthetic [21]. A set of samples from UTSig dataset are shown in TABLE IV. More information about the used datasets is available in TABLE V.

The signature images from the datasets need to be first pre-processed before the feature learning step. We followed the preprocessing approach suggested in [6] which included removing the background by OTSU's algorithm [22], inverting the image brightness, and resizing to the input size of the network. As aforementioned, we used deep residual networks in this research, with the input size of 242x242 pixels. The details of our used ResNet models are thoroughly described in TABLE III and Fig. 2.

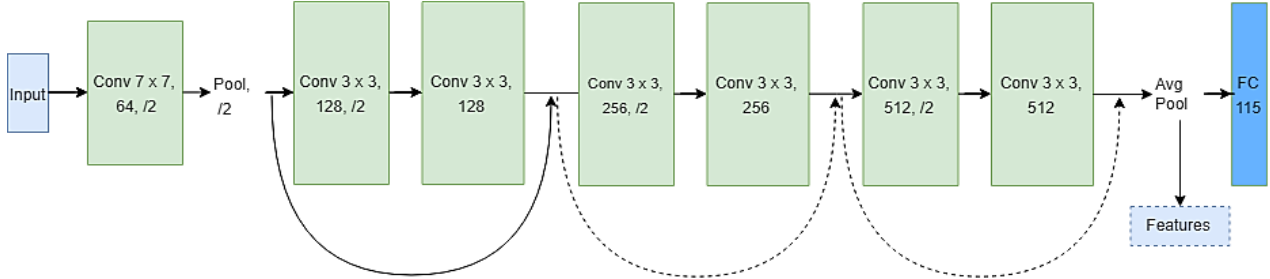

Fig. 2 The CNN architecture that was used in this research. The input first goes through a convolutional layer with a max pooling layer followed by. As it is was mentioned before, the CNN consists of three residual blocks that each have two convolutional layers. Each block adds its input to its last layer output and projects it as the final output of the block. Lastly, the final block's output will be manipulated by an average pooling activation function. The output of this sequence is the feature vector that will be used in our OSV task.

As it was stated before, for each user there was a classifier that was trained on a two-class problem between genuine and random forgery signatures of the particular signature class. We used Scikit-Learn python package to create and train SVMs with the different kernels that were mentioned above. The parameters used for the SVMs were set as the default parameters of the SVC module, available in the library. Additionally, the SVM balance factor was also set in order to handle unbalanced classifications. Based on this factor, the adaptation algorithm automatically adjusts weights inversely proportional between the sizes of two classes[23].

The mentioned protocol was used to set the parameters of the proposed algorithm for OSV. We then used 5, 7, and 10 genuine signatures of each dataset for the training phase of both feature generator and the SVM classifier of each user, as it can be seen in TABLE VI, TABLE VII, and TABLE VIII.

For the SVM classifiers, a number of 200 random forgeries were generated. The performance of the proposed method was reported based on the Equal Error Rate (EER). In the testing phase, we used the skilled forgery samples alongside a pool of unseen genuine samples of each class to obtain EERs of the network. All error rates were obtained by averaging five different runs. We also experimented with different SVM kernels for each dataset, which can be seen in [24]. The source task in these experiments has always been writer identification.

### A. Comparison with the state-of-the-art in different signature datasets

The state-of-the-art performances on the three signature datasets are compared with our proposed method in this part. Tables IX-XI present the comparison with the state-of-the-art performance on UT-Sig, GPDS-Synthetic, and MCYT, respectively. On UT-Sig, we achieved EER of 9.80% which is substantially better than the best reported EER of 17.45% by [7] in the literature. On MCYT, there were many studies on this data in which the recent study [6] reported very good accuracies, as shown in TABLE XI. We achieved EER of 3.98% which is comparable to the best EER of 2.87% in the literature. Putting it all together, the results confirm that our proposed transfer learning approach could achieve the state-of-the-art results on GPDS-Synthetic and UTSig datasets and also comparable performance on MCYT dataset.

TABLE II   DIFFERENT KERNELS THAT WERE USED IN THIS EXPERIMENT. LOG KERNEL WAS THE MAIN KERNEL THAT WAS USED IN THE EXPERIMENTS.

| Cosine Distance Kernel | $k(x,y) = \dfrac{xy^T}{\|\|x\|\|\|\|y\|\|}$ |
|---|---|
| Log Kernel | $k(x,y) = -\log\left(\|\|x-y\|\|^d + 1\right)$ |
| Linear Kernel | $k(x,y) = xy^T$ |
| RBF Kernel | $k(x,y) = e^{-\gamma\|\|x-y\|\|^2}$ |

TABLE III. PROPERTIES OF THE DEEP NETWORK. RESIDUAL BLOCKS ARE THREE TIMES REPEATED. N = 1, 2, 3.

| Layer | Properties |
|---|---|
| Input | 242*242 input size |
| Conv | 64, K = 7 * 7, stride=2 |
| Batch Norm | - |
| Relu | - |
| Avg Pool | K = 3 * 3, stride = 2 |
| Conv | N*128, K = 3 * 3, stride = 2 |
| Batch Norm | - |
| Relu | - |
| Conv | N*128, K = 3 * 3, stride = 1 |
| Batch Norm | - |
| Add | - |
| Avg Pool | - |
| **Fully Connected** | - |

TABLE IV. A SAMPLE POOL OF UTSIG. AS IT IS APPARENT PERSIAN SIGNATURES HAVE UNIQUE CHARACTERISTICS COMPARED TO LATIN SIGNATURES.

| Genuine | Average | Skilled Forgery |
|---|---|---|
| | | |
| | | |

TABLE V. SIGNATURE DATASETS PROPERTIES

| Dataset | Nationality | #users and #images | #Genuine signatures | #Skilled forgeries |
|---|---|---|---|---|
| UT-Sig | Persian | 115 users with 8280 images | 27 | 42 |
| MCYT-75 | Spanish | 75 users with 2250 images | 15 | 15 |
| GPDS-synthetic | Synthesized signatures | 4000 users with 216000 images | 24 | 30 |

TABLE VI. THE PERFORMANCE (EER %) OF THE MODEL ON UTSIG. The first column shows the number of used training signatures. The second column includes two sub-columns: Word Recognition (Rec) and with fine-tuning. The Word Rec. column refers to the results of feature learning by the CNN trained with word recognition task while the right column includes the additional fine-tuning by the signature images. The third column also includes two columns: Writer Identification (ID.) and with fine-tuning. The Writer ID column refers to the results of feature learning by the CNN trained with writer identification task while the right column includes the additional fine-tuning by the signature images. The last column on the right shows the results of training from scratch without using the pre-trained CNN on handwriting. The lowest EER in each row is presented boldfaced.

| Number of Training Signatures per User | Word Rec. | Word Rec. with fine-tuning | Writer ID | Writer ID. with fine-tuning | Training from scratch |
|---|---|---|---|---|---|
| 5 samples | 13.6% | 11.92% | 15.31% | **11.26%** | 12.18% |
| 7 samples | 13.00% | 11.35% | 14.65% | **9.94%** | 11.46% |
| 10 samples | 11.33% | 9.93% | 10.60% | **9.02%** | 9.54% |

TABLE VII. THE PERFORMANCE (EER) OF THE MODEL ON MCYT-75

| Number of Training Signatures | Word Rec. | Word Rec. With fine-tuning | Writer ID. | Writer ID. with fine-tuning | Training from scratch |
|---|---|---|---|---|---|
| 5 samples | 9.18% | 7.92% | 8.83% | **7.12%** | 8.09% |
| 7 samples | 8.05% | 5.78% | 7.32% | **5.51%** | 5.58% |
| 10 samples | 6.16% | 4.67% | 5.94% | **3.98%** | 4.42% |

TABLE VIII. THE PERFORMANCE (EER) OF THE MODEL ON GPDS-SYNTHETIC

| Number of Training signatures | Word Rec. | Word Rec. with fine-tuning | Writer ID. | Writer ID. with fine-tuning | Training from scratch |
|---|---|---|---|---|---|
| 5 samples | 14.07% | 8.17% | 15.09% | **7.99%** | 8.42% |
| 7 samples | 14.46% | 7.51% | 13.54% | **7.34%** | 8.11% |
| 10 samples | 10.07% | 6.99% | 11.99% | **6.81%** | 7.04% |

TABLE IX. DIFFERENT SVM KERNELS PERFORMANCE (EER)

| | Datasets | Log Kernel | Linear Kernel | Poly Kernel | RBF Kernel |
|---|---|---|---|---|---|
| 10 Genuine Signatures Used per Each User | UTSig | **10.08%** | 14.77% | 13.31% | 11.08% |
| | MCYT-75 | **5.18%** | 9.99% | 6.07% | 5.60% |
| | GPDS-Synthetic | **9.07%** | 12.98% | 11.78% | 11.89% |

TABLE X. COMPARISON WITH THE STATE-OF-THE-ART IN UTSIG

| Reference | Features | Classifier | WD or WI | # Genuine Samples | EER (%) |
|---|---|---|---|---|---|
| [7] | DRT+DMML | Thresholding | WD | 12 | 20.28 |
| [7] | HOG+DMML | Thresholding | WD | 12 | 17.45 |
| [20] | Fixed-point geometrics | SVM | WD | 12 | 29.71 |
| Our Study | Transfer Learning from Handwriting | SVM | WD | 5 | **11.16** |
| | | | | 7 | **9.96** |
| | | | | 10 | **9.80** |

TABLE XI. COMPARISON WITH THE STATE-OF-THE-ART IN GPDS-SYNTHETIC.

| Reference | Feature | Classifier | WD or WI | # Genuine Samples | # Users | EER |
|---|---|---|---|---|---|---|
| [7] | HOG + DMML | Thresholding | WD | 10 | 2500 | 12.80 |
| | | Thresholding | WD | 10 | 4000 | 13.30 |
| [21] | LBP | SVM | WI | 10 | 4000 | 16.44 |
| [25] | HOT | AIRSV | WD | 10 | 4000 | 16.68 |
| Our Study | Transfer Learning from Handwriting | SVM | WD | 5 | 4000 | **7.99** |
| | | | | 7 | 4000 | **7.34** |
| | | | | 10 | 4000 | **6.81** |

TABLE XII. COMPARISON WITH STATE-OF-THE-ART IN MCYT-75

| Reference | Feature | Classifier | WD or WI | #Genuine Samples | EER (%) |
|---|---|---|---|---|---|
| [26] | DRT+PCA | PNN | WD | 10 | 9.87 |
| [7] | HOG + DMML | Thresholding | WD | 10 | 9.86 |
| [6] | SigNet | SVM | WD | 5 | 3.58 |
| | | | | 10 | **2.87** |
| [27] | Global + local features | SVM | WD | 5 | 9.16 |
| | | | | 10 | 7.92 |
| Our Study | Transfer Learning from Handwriting | SVM | WD | 5 | **7.12** |
| | | | | 7 | **5.51** |
| | | | | 10 | **3.98** |

**REFERENCES**

[1] L. G. Hafemann, R. Sabourin, and L. S. Oliveira, "Offline handwritten signature verification-literature review," *arXiv preprint arXiv:1507.07909,* 2015.

[2] S. Jabin and F. J. Zareen, "Biometric signature verification," *International Journal of Biometrics,* vol. 7, no. 2, pp. 97-118, 2015.

[3] B. Shekar, R. Bharathi, J. Kittler, Y. V. Vizilter, and L. Mestestskiy, "Grid structured morphological pattern spectrum for off-line signature verification," in *Biometrics (ICB), 2015 International Conference on*, 2015, pp. 430-435: IEEE.

[4] J. Hu, Z. Guo, Z. Fan, and Y. Chen, "Offline Signature Verification Using Local Features and Decision Trees," *International Journal of Pattern Recognition and Artificial Intelligence,* vol. 31, no. 03, p. 1753001, 2017.

[5] V. Noroozi, L. Zheng, S. Bahaadini, S. Xie, and P. S. Yu, "SEVEN: Deep Semi-supervised Verification Networks," *arXiv preprint arXiv:1706.03692,* 2017.

[6] L. G. Hafemann, R. Sabourin, and L. S. Oliveira, "Learning features for offline handwritten signature verification using deep convolutional neural networks," *Pattern Recognition,* vol. 70, pp. 163-176, 2017.

[7] A. Soleimani, B. N. Araabi, and K. Fouladi, "Deep Multitask Metric Learning for Offline Signature Verification," *Pattern Recognition Letters,* vol. 80, pp. 84-90, 2016.

[8] F. Vargas, M. Ferrer, C. Travieso, and J. Alonso, "Off-line handwritten signature GPDS-960 corpus," in *Document Analysis and Recognition, 2007. ICDAR 2007. Ninth International Conference on*, 2007, vol. 2, pp. 764-768: IEEE.

[9] X. Shu, G.-J. Qi, J. Tang, and J. Wang, "Weakly-shared deep transfer networks for heterogeneous-domain knowledge propagation," in *Proceedings of the 23rd ACM international conference on Multimedia*, 2015, pp. 35-44: ACM.

[10] Y. Zhu *et al.*, "Heterogeneous Transfer Learning for Image Classification," in *AAAI*, 2011.

[11] W. Dai, Y. Chen, G.-R. Xue, Q. Yang, and Y. Yu, "Translated learning: Transfer learning across different feature spaces," in *Advances in neural information processing systems*, 2009, pp. 353-360.

[12] D. C. Cireşan, U. Meier, and J. Schmidhuber, "Transfer learning for Latin and Chinese characters with deep neural networks," in *Neural Networks (IJCNN), The 2012 International Joint Conference on*, 2012, pp. 1-6: IEEE.

[13] R. R. Nair, N. Sankaran, B. U. Kota, S. Tulyakov, S. Setlur, and V. Govindaraju, "Knowledge transfer using Neural network based approach for handwritten text recognition," in *2018 13th IAPR International Workshop on Document Analysis Systems (DAS)*, 2018, pp. 441-446: IEEE.

[14] U.-V. Marti and H. Bunke, "The IAM-database: an English sentence database for offline handwriting recognition," *International Journal on Document Analysis and Recognition,* vol. 5, no. 1, pp. 39-46, 2002.

[15] G. Alvarez, B. Sheffer, and M. Bryant, "Offline Signature Verification with Convolutional Neural Networks," Tech. rep., Stanford University, Stanford2016.

[16] K. He, X. Zhang, S. Ren, and J. Sun, "Deep residual learning for image recognition," in *Proceedings of the IEEE conference on computer vision and pattern recognition*, 2016, pp. 770-778.

[17] J. Sadri, M. R. Yeganehzad, and J. J. P. R. Saghi, "A novel comprehensive database for offline Persian handwriting recognition," vol. 60, pp. 378-393, 2016.

[18] C. Cortes and V. J. M. L. Vapnik, "Support-vector networks," journal article vol. 20, no. 3, pp. 273-297, September 01 1995.

[19] J. Fierrez-Aguilar, N. Alonso-Hermira, G. Moreno-Marquez, and J. Ortega-Garcia, "An off-line signature verification system based on fusion of local and global information," *Lecture notes in computer science,* pp. 295-306, 2004.

[20] A. Soleimani, K. Fouladi, and B. N. Araabi, "UTSig: A Persian offline signature dataset," *IET Biometrics,* vol. 6, no. 1, pp. 1-8, 2017.

[21] M. A. Ferrer, M. Diaz-Cabrera, and A. Morales, "Static signature synthesis: A neuromotor inspired approach for biometrics," *IEEE Transactions on Pattern Analysis and Machine Intelligence,* vol. 37, no. 3, pp. 667-680, 2015.

[22] N. Otsu, "A threshold selection method from gray-level histograms," *IEEE transactions on systems, man, and cybernetics,* vol. 9, no. 1, pp. 62-66, 1979.

[23] "<DOES DIVERSITY IMPROVE DEEP LEARNING..pdf>."

[24] A. Patle and D. S. Chouhan, "SVM kernel functions for classification," in *Advances in Technology and Engineering (ICATE), 2013 International Conference on*, 2013, pp. 1-9: IEEE.

[25] Y. Serdouk, H. Nemmour, and Y. Chibani, "Handwritten signature verification using the quad-tree histogram of templates and a Support Vector-based artificial immune classification," *Image and Vision Computing,* vol. 66, pp. 26-35, 2017.

[26] S. Y. Ooi, A. B. J. Teoh, Y. H. Pang, and B. Y. Hiew, "Image-based handwritten signature verification using hybrid methods of discrete radon transform, principal component analysis and probabilistic neural network," *Applied Soft Computing,* vol. 40, pp. 274-282, 2016.

[27] M. Sharif, M. A. Khan, M. Faisal, M. Yasmin, and S. L. Fernandes, "A framework for offline signature verification system: Best features selection approach," *Pattern Recognition Letters,* 2018.